\pdfoutput=1

\documentclass[11pt]{article}

\usepackage{acl}

\usepackage{times}
\usepackage{latexsym}

\usepackage[T1]{fontenc}

\usepackage[utf8]{inputenc}

\usepackage{microtype}

\usepackage{graphicx}
\usepackage[inline]{enumitem}
\usepackage{multirow}
\usepackage{amsmath}
\usepackage{hyperref}
\usepackage{cleveref}[2012/02/15]
\usepackage{pifont}
\usepackage{pgfplots}
\usetikzlibrary{patterns}

\pgfplotsset{compat=1.17}

\crefformat{footnote}{#2\footnotemark[#1]#3}

\newcommand{\enhok}[0]{En$\rightarrow$Hokkien}
\newcommand{\hoken}[0]{Hokkien$\rightarrow$En}
\newcommand{\enhokbi}[0]{En$\leftrightarrow$Hokkien}
\newcommand{\enzhbi}[0]{En$\leftrightarrow$Zh}
\newcommand{\hok}[0]{Hokkien}
\newcommand{\tailo}[0]{T\^{a}i-l\^{o}}
\newcommand{\cmark}{\ding{51}}
\newcommand{\xmark}{\ding{55}}

\title{Speech-to-Speech Translation For A Real-world Unwritten Language}

\author{Peng-Jen Chen, Kevin Tran, Yilin Yang, Jingfei Du, Justine Kao
 \\ {\bf Yu-An Chung, Paden Tomasello, Paul-Ambroise Duquenne}
 \\ {\bf Holger Schwenk, Hongyu Gong, Hirofumi Inaguma, Sravya Popuri}
 \\ {\bf Changhan Wang, Juan Pino, Wei-Ning Hsu, Ann Lee} \\
  {Meta AI} \\
  \texttt{\{pipibjc,annl\}@meta.com} \\
}

\begin{document}
\maketitle
\begin{abstract}
We study speech-to-speech translation (S2ST) that translates speech from one language into another language and focuses on building systems to support languages without standard text writing systems. We use English-Taiwanese Hokkien as a case study, and present an end-to-end solution from training data collection, modeling choices to benchmark dataset release. First, we present efforts on creating human annotated data, automatically mining data from large unlabeled speech datasets, and adopting pseudo-labeling to produce weakly supervised data. On the modeling, we take advantage of recent advances in applying self-supervised discrete representations as target for prediction in S2ST and show the effectiveness of leveraging additional text supervision from Mandarin, a language similar to Hokkien, in model training. Finally, we release an S2ST benchmark set to facilitate future research in this field\footnote{The demo can be found at \url{https://huggingface.co/spaces/facebook/Hokkien_Translation}.}.
\end{abstract}

\section{Introduction}
Speech-to-speech translation (S2ST) aims at translating speech from one language into speech in another language. S2ST technology can not only enable communication between people speaking different languages but also help knowledge sharing across the world.
Conventionally, S2ST can be achieved via the concatenation of three systems: automatic speech recognition (ASR), machine translation (MT) and text-to-speech synthesis (TTS)~\citep{lavie1997janus,nakamura2006atr}.
In recent years, the advancement from end-to-end speech-to-text translation (S2T)~\citep{berard2016listen} or text-to-speech translation (T2ST)~\citep{zhang2020uwspeech,lee2022direct} have simplified the S2ST pipeline into two stages, which reduces error propagation issues and improves efficiency~\cite{lee2022direct}.
Most recently, researchers have built one-stage S2ST systems~\citep{jia2019direct} that jointly optimize intermediate text generation and target speech generation steps~\citep{kano2021transformer,jia2022translatotron,inaguma2022unity} or further remove the dependency on text completely~\citep{tjandra2019speech,lee2022direct,lee2021textless}.
Directly conditioning on the source speech during the generation process allows the systems to transfer non-linguistic information, such as speaker voice, from the source directly~\citep{jia2022translatotron}. Not relying on text generation as an intermediate step allows the systems to support translation into languages that do not have standard or widely used text writing systems~\citep{tjandra2019speech,zhang2020uwspeech,lee2021textless}.

While more than 40\% of the languages in the world do not have text written forms\footnote{\url{https://www.ethnologue.com}}, S2ST for unwritten languages still remains a research area with little exploration mainly due to the lack of training data.
The majority of the previous work on this topic conducts experiments on datasets built from applying TTS on S2T corpora to generate synthetic target speech for model training~\citep{tjandra2019speech,zhang2020uwspeech}.
\citet{lee2021textless} presents the first textless S2ST system trained on real S2ST data, while it only investigates translation between high-resource and similar language pairs (English$\leftrightarrow$Spanish, English$\leftrightarrow$French).
The feasibility of S2ST for unwritten languages under a low-resource setup, which is a more realistic scenario, remains unknown.

In this work, we take Taiwanese \hok~as an example of an unwritten language and study S2ST between English (En) and Taiwanese \hok.
Taiwanese \hok~(hereafter \hok) is one of the official languages in Taiwan spoken by over 70\% of the population (approximately 15.8 million people).
As a language that is passed down predominantly orally, \hok~lacks a unitary writing system that is widely adopted by its native speakers, though a few possible writing systems exist, e.g.~Chinese characters (Hanji), or romanization systems such as Peh-\={o}e-j\={\i} (POJ) and \tailo, etc. 
In addition,~\hok~is a tonal language that has complex tone sandhi rules~\citep{cheng1968tone}.
\citet{wang2004multiple} investigates Mandarin-Taiwanese~\hok~S2ST with a cascaded template matching approach.
In our work, we focus on~\enhokbi, a distant language pair, and build one-stage S2ST systems.

We take advantage of the discrete unit-based S2ST approach~\citep{lee2022direct}, which applies a self-supervised speech encoder to convert the target speech into a sequence of integers and translates source speech into target discrete units, to build the \enhokbi~systems.
First, to support \enhok~translation, we extend HuBERT-based discrete unit extraction~\citep{hsu2021hubert} and examine the feasibility of unit-to-waveform generation~\citep{polyak2021speech} for tonal languages.
Second, we leverage the unit-based speech normalization technique proposed in~\citet{lee2021textless} to remove the non-linguistic variations in speech from multiple speakers.
The original study takes advantage of synthetic speech generated from TTS as the reference target for normalization, while we build the normalizer with real \hok~speech data.
Last but not least, we study two S2ST model training strategies, speech-to-unit translation (S2UT) with a single decoder~\citep{lee2022direct} or a two-pass decoding process~\citep{inaguma2022unity} that leverages Mandarin (Zh) as a written language similar to \hok~to provide extra text supervision.

As no S2ST dataset is available for \enhokbi, we also leverage Mandarin to assist the parallel S2ST data creation process and create a 60-hr human annotated training set and an open benchmark set.
Nevertheless, this is still a low-resource problem and to tackle the data scarcity issue, we further apply En$\leftrightarrow$Zh MT to create weakly supervised data~\citep{popuri2022enhanced,dong2022leveraging} and learn a joint embedding space for English and \hok~through Mandarin to support automatic data mining from unlabeled English and \hok~data~\citep{duquenne2021multimodal}.

The contributions of this work are as follows:
\begin{itemize}
    \item We present empirical studies that consolidate various state-of-the-art techniques for S2ST that were previously studied in a controlled setup with synthetic speech and verify their effectiveness in \enhokbi~translation, where \hok~is a language without a widely adopted standard text writing system.
    \item A benchmark set on \enhokbi~S2ST and the evaluation model for \hok~speech is released \footnote{\url{https://sites.google.com/nycu.edu.tw/speechlabx/tat_s2st_benchmark}} to encourage future research in this direction.
    \item To the best of our knowledge, we are the first to build one-stage S2ST systems for an unwritten language in a real-world scenario.
\end{itemize}

\section{Related Work}

Existing S2ST models can be categorized in several aspects. First,~\citet{jia2019direct,jia2022leveraging,jia2022translatotron} directly predict spectrogram as the model output, 
while~\citet{lee2022direct,lee2021textless,huang2022transpeech, popuri2022enhanced,inaguma2022unity} leverage self-supervised speech model such as HuBERT~\citep{hsu2021hubert} to encode the target speech into a sequence of discrete units and apply knowledge from speech-to-text modeling to S2ST.
Second,~\citet{jia2019direct,jia2022translatotron} require extra supervision from target text or phonemes during model training, 
while~\citet{tjandra2019speech,lee2021textless,popuri2022enhanced} show the possibility of model training with speech data only.
Finally,
~\citet{kano2021transformer,inaguma2022unity} concatenate multiple decoders learned with additional text targets or speech units with different granularity and perform multi-pass decoding during inference.

While the modeling choices may vary, one common challenge for S2ST model training is the data scarcity issue.
\citet{jia2022cvss} applies high-quality English TTS and creates an X$\rightarrow$En S2ST dataset with synthetic target speech for 21 languages.
To create S2ST datasets with real speech,~\citet{wang-etal-2021-voxpopuli} aligns ASR transcripts for more than 100 language pairs, and \citet{duquenne2022speechmatrix} applies distance-based bitext mining to audio, producing a mined S2ST dataset between 17 European languages.
Weakly supervised data created from TTS~\citep{jia2022leveraging} or a cascaded pipeline with ASR and MT models~\citep{dong2022leveraging,popuri2022enhanced} is often combined with the parallel S2ST data.
In addition, self-supervised pre-training with large-scale unlabeled data is also effective in improving S2ST model performance~\citep{jia2022leveraging,popuri2022enhanced}.

\section{Methods}
\begin{figure*}[ht!]
  \centering
  \includegraphics[width=0.95\linewidth]{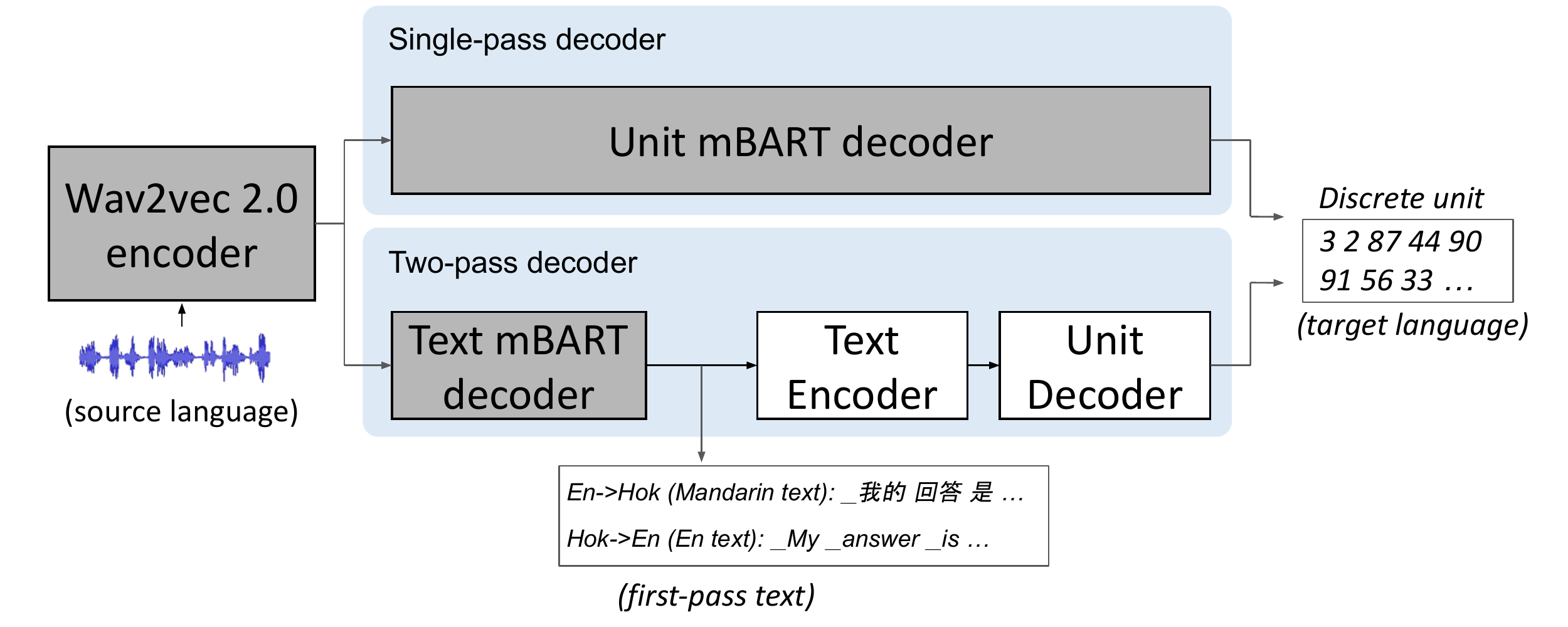}
  \caption{Model architecture of S2ST with single-pass and two-pass decoder. The blocks in shade illustrate the modules that are pre-trained. Text in italic is the training objective. 
  }
  \label{fig:flowchart}
  \vspace{-0.5cm}
\end{figure*}

In this section, we first present two types of backbone architectures for S2ST modeling.
Then, we describe our efforts on creating parallel S2ST training data from human annotations as well as leveraging speech data mining~\citep{duquenne2021multimodal} and creating weakly supervised data through pseudo-labeling~\citep{popuri2022enhanced,jia2022leveraging}.

\subsection{Model architectures}
\label{sec:s2st_arch}
As illustrated in Fig.~\ref{fig:flowchart}, we study one model architecture that applies a single-pass decoding process and directly translates source speech to the target, and the second one relies on target text (Mandarin text in the case of \hok~speech) to provide extra supervision and performs two-pass decoding.
Both architectures predict discrete units as the target, and the speech encoder and text or unit decoders are pre-trained with unlabeled speech or text data.

\subsubsection{Speech-to-unit translation (S2UT)}
We follow the S2UT approach proposed in~\citet{lee2022direct} and adopt HuBERT~\citep{hsu2021hubert} to convert target speech into discrete units via k-means on intermediate representation.
While \hoken~systems can be trained on target English speech generated from single-speaker TTS to remove variations in accents from multiple speakers or noises from different recording conditions,
when training \enhok~systems, we first apply a unit-based speech normalizer~\citep{lee2021textless} on the real \hok~target speech.
The speech normalizer is built by applying Connectionist Temporal Classification (CTC)~\citep{graves2006connectionist} finetuning with the \hok~HuBERT model using multi-speaker speech as input and the corresponding discrete units extracted from real \hok~speech from a reference speaker as target.

The resulting S2ST system consists of a sequence-to-sequence S2UT model and a unit-based HiFi-GAN vocoder~\citep{polyak2021speech} for unit-to-waveform conversion.
The original S2UT model design in~\citet{lee2022direct} consists of a speech encoder and a discrete unit decoder.
\citet{inaguma2022unity} includes a stack of an intermediate text decoder and encoder before the unit decoder to take advantage of additional supervision from target text to improve model performance and carry out two-pass decoding.
We describe the differences in the decoding process of the two model designs in the following sections.

For both models, we pre-train the speech encoder with Conformer-based~\citep{gulati2020conformer} wav2vec~2.0~\citep{baevski2020wav2vec,popuri2022enhanced} using a large amount of unlabeled speech.
To speed up model training, we remove the multi-layer convolutional feature encoder that converts input waveform to latent speech representations at every 20-ms, and take in pre-computed 80-dimensional log-mel filterbank features stacked at every two frames instead.
Preliminary experiments show no performance degradation with filterbank input.

\subsubsection{Single-pass decoding S2UT}
\citet{lee2022direct} proposes to use a single unit decoder, which can be trained with standard cross-entropy loss.
Following~\citet{popuri2022enhanced}, we apply mBART training~\citep{liu2020multilingual}, which is a sequence-to-sequence autoencoder trained with a denoising objective across monolingual text in multiple languages, using discrete units extracted from unlabeled speech with consecutive duplicate units removed, and use the pre-trained decoder to initialize the unit decoder.
During decoding, we perform single-pass beam search with the unit decoder.

\subsubsection{Two-pass decoding S2UT: UnitY}
UnitY model~\citep{inaguma2022unity} also performs speech-to-unit translation, while it includes a target text decoder and a target text encoder before the unit decoder and incorporates target text prediction as an auxiliary loss during training.
All the modules are trained jointly.
When training the \enhok~model, we use Mandarin as the target text due to its proximity to \hok~and abundance in text data.
The availability of Mandarin-\hok~bilingual population also allows us to create human annotated training data described in the next section.

We follow~\citet{inaguma2022unity} to apply R-Drop~\citep{wu2021rdrop} regularization during training as well as initializing the target text decoder with a text mBART model~\citep{liu2020multilingual} pre-trained on the combination of En and Zh monolingual text data.

\subsection{Training data}
In the following sections, we describe three different efforts on creating parallel \enhokbi~data for model training.

\subsubsection{Supervised human annotated data}
\label{sec:method_human_data}
Since there are not many \enhokbi~bilingual speakers who can directly translate between the two languages, we use Mandarin as a pivot language during the data creation process whenever possible. 
We sample from the following data sources and adopt different strategies to create human annotated parallel data:
\begin{enumerate*}[label=(\arabic*)]
  \item \hok~dramas, which include \hok~speech and aligned Mandarin subtitles\footnote{\hok~drama data is obtained from the collaboration with National Taiwan University.},
  \item Taiwanese Across Taiwan (TAT)~\citep{liao2020formosa}, a \hok~read speech dataset containing transcripts in \tailo~and Hanji, and
  \item MuST-C v1.2 En-Zh S2T data~\citep{cattoni2021must}.
\end{enumerate*}

We ask Zh-En bilinguals to translate the subtitles of the \hok~dramas into English to create \hoken~S2T data.
For the TAT dataset, we leverage a small group of \enhokbi~bilinguals, who have access to both the \hok~audio and the transcripts, to translate the \hok~speech directly into English text.

For MuST-C, we ask Zh-\hok~bilinguals to translate the Mandarin text into a mix of \tailo~and Hanji and then record the script into \hok~speech\footnote{The annotators pointed out that it is easier to leverage both systems, which is another evidence of \hok~lacking a commonly adopted text writing system.}.
We found that having an intermediate non-standardized script was still helpful in improving the fluency and accuracy of the created \hok~speech, as opposed to asking speakers to do Mandarin text-to-\hok~speech translation on the fly, while no \tailo~or Hanji transcripts are used during \enhok~S2ST system training.

In the end, we build an \enhok~S2ST training set from MuST-C.
For \hoken~training, we apply an English text-to-unit (T2U) model~\citep{lee2021textless}, which is a sequence-to-sequence Transformer model trained on English characters as input and units extracted from the corresponding speech as target, on the English text collected for \hok~dramas and TAT, as well as the English transcriptions provided in MuST-C, to convert the text into units.

\subsubsection{Mined data}


To build a shared embedding space for \hok~and English speech and text data for performing speech-to-text or speech-to-speech mining at scale, we again take advantage of Mandarin text as the bridge between the two languages.
First, to encode En and Zh text in the same embedding space, we apply the method proposed in~\citet{duquenne2022t} to finetune \texttt{XLM-R LARGE}~\citep{conneau2019cross} to fit LASER~\citep{Artetxe:2018:tacl_laser} English text space using Zh-En parallel MT data.
Then, we minimize the mean squared error (MSE) loss between the max-pooled output of the learned text encoder and that of a speech encoder using aligned \hok~speech and Mandarin or English text\footnote{A subset of the \hok~dramas data has English subtitles.}.
The text encoder is fixed during speech encoder training, where the latter is initialized with Conformer-based wav2vec 2.0 pre-trained with \hok~speech, and this process further encodes the \hok~speech, Mandarin and English text in the same embedding space.
Similarly, we also train an En speech encoder using speech and text pairs from En ASR data.
In the end, we create a shared embedding space for En speech and text, Mandarin text, and Hokkien speech, which supports En text and \hok~speech or En speech and \hok~speech mining based on cosine similarity.


\subsubsection{Weakly supervised data}
We take advantage of cascaded systems to create weakly supervised data from monolingual speech data~\citep{popuri2022enhanced,dong2022leveraging}.
For~\enhok, we leverage En ASR data and apply En$\rightarrow$Zh MT on the transcriptions, followed by a Zh$\rightarrow$\hok~text-to-unit-translation (T2UT) model, which is a Transformer-based sequence-to-sequence model trained with character encoded Mandarin text as input and normalized units encoded from corresponding \hok~speech as targets.
For \hoken, we apply the Zh$\rightarrow$En MT model on the \hok~drama Mandarin subtitle, followed by En T2U~\citep{lee2021textless} to create pseudo-labeled data.

\section{Experimental Setup}
In this section, we describe the data, model training details, as well as baseline systems and the evaluation protocol. 
All experiments are conducted using \texttt{fairseq}~\citep{ott2019fairseq}.

\subsection{Data}

\begin{table}
\centering
\resizebox{\linewidth}{!}{
\begin{tabular}{l|c|cc}
\hline
  & \multirow{2}{*}{Data source} & Source & Target \\
  & & speech (hrs) & speech (hrs) \\
\hline
\multirow{3}{*}{\hoken} & \hok~dramas & 5.8$^\ast$ & synthetic \\
\cline{2-4}
 & TAT & 4.6 (74M, 86F) & synthetic  \\
\cline{2-4}
 & \multirow{2}{*}{MuST-C} & 51 (8M, 14F) & synthetic \\
\cline{1-1}
\cline{3-4}
\enhok & & 35$^\ast$ & 51 (8M, 14F) \\
\hline
\end{tabular}}
\caption{Statistics of the human annotated training sets. (M: male, F: female, $^\ast$: no gender information available)}
\label{tab:training_stats}
\end{table}

\begin{table}
\centering
\resizebox{\linewidth}{!}{
\begin{tabular}{l|c|ccc}
\hline
 &  & \# samples & Duration (hrs) & \# speakers \\
\hline
\multirow{2}{*}{Dev} & En & \multirow{2}{*}{722} & 1.62 & 10 (5 M, 5 F)  \\
 & \hok & & 1.46 & 10 (8 M, 2 F) \\
\hline
\multirow{2}{*}{Test} & En & \multirow{2}{*}{686} & 1.47 & 10 (5 M, 5 F)  \\
 & \hok & & 1.42 & 10 (3 M, 7 F)\\
\hline
\end{tabular}}
\caption{Statistics of the TAT-S2ST benchmark set. (M: male, F: female)}
\label{tab:tat_s2st_stats}
\end{table}

\subsubsection{Supervised human annotated data}
\label{sec:human_annotated_data}
We carry out the annotation process in Sec.~\ref{sec:method_human_data}, and Table~\ref{tab:training_stats} summarizes the statistics of the training data.
In the end, we create a 61.4-hr human annotated training set for \hoken, and 35-hr for \enhok.
We do not combine the synthetic English speech created from English text translation with the real \enhok~S2ST dataset during training.

\subsubsection{TAT-S2ST: \enhokbi~S2ST evaluation dataset}
\label{sec:tat_s2st_eval_set}
As a part of the effort on creating human annotated data, we also create an \enhokbi~S2ST benchmark set 
to facilitate future research in the field.
The English text translation we collect for the TAT dev and test sets are proofread and corrected for fluency and naturalness.
We then recruit native speakers to read out loud and record the English text translations, producing \enhokbi~parallel speech data.
Table~\ref{tab:tat_s2st_stats} shows the statistics of this benchmark set.
While \hok~does not have a standardized and widely adopted writing system, TAT provides \tailo~transcripts, which is a romanization system for \hok, for the \hok~audios, which can be leveraged as reference text in evaluation (Sec.~\ref{sec:evaluation}).

\subsubsection{Mined data}
\label{sec:exp_setup_mined_data}
We train the En and Zh joint text encoder on CCMatrix~\citep{schwenk2019ccmatrix}, the \hok~speech encoder on \hok~dramas, and the English speech encoder on English ASR data from CommonVoice~\citep{DBLP:conf/lrec/ArdilaBDKMHMSTW20}, CoVoST-2~\citep{wang2021covost}, Europarl-ST~\citep{DBLP:conf/icassp/Iranzo-SanchezS20}, MuST-C~\citep{di-gangi-etal-2019-must}, Voxpopuli~\citep{wang-etal-2021-voxpopuli} and Librispeech~\citep{panayotov2015librispeech}.
The learning rate is set to $10^{-4}$, with an inverse square root schedule. The maximum number of tokens is set to 640k (equivalent to 40 seconds with 16kHz sampling rate), with a maximum number of sentences set to 32.
We train the models with 48 GPUs for 60k steps. 

With the trained text and speech encoders, we perform data mining between \hok~speech from \hok~dramas and English Common Crawl text, and between the former and Librivox English audio\footnote{\url{https://librivox.org/api/}}.
We post-process the mined data in order to have a maximum of 20\% overlap between any two audio segments.
In the end, we obtain 8.1k-hr~\hoken~S2T mined data and 197-hr \enhokbi~S2ST mined data, respectively.
The difference in the volume is mainly due to the domain mismatch in audiobooks from Librivox and \hok~dramas.

\subsubsection{Weakly supervised data}
\label{sec:weakly_supervised_data}
For \enhok, we apply En$\rightarrow$Zh MT on the combination of the English transcripts from Librispeech~\citep{panayotov2015librispeech} and TED-LIUM3~\citep{hernandez2018ted}, totaling 1.5k-hr of English speech.
The En$\rightarrow$Zh MT model is a 12-layer Transformer model trained on CCMatrix~\citep{schwenk2019ccmatrix} using disjoint BPEs for En and Zh encoded by the
\texttt{sentencepiece} toolkit~\citep{kudo2018sentencepiece}, each of size 32768. We use 16 GPUs, a batch size of 14,336 tokens and a learning rate of $10^{-3}$ during training.

The Zh$\rightarrow$\hok~T2UT model following the En$\rightarrow$Zh translation step is trained on \hok~dramas and the aligned Mandarin subtitles.
We filter out speech containing Mandarin code-switching by applying Mandarin ASR and computing the Levenshtein distance between the ASR output and the subtitles, as well as short sentences with less than three characters, resulting in 1k-hr \hok~speech for training.

For \hoken, we apply Zh$\rightarrow$En MT on the Mandarin subtitles from 8k-hr \hok~drama data, followed by an En T2U trained on LJSpeech~\citep{ljspeech17} used in~\citet{lee2021textless}.
We leverage the open-sourced NLLB-200 3.3B-parameter multilingual model\footnote{\url{https://github.com/facebookresearch/fairseq/tree/nllb/examples/nllb/modeling}}~\citep{costa2022no} for Zh$\rightarrow$En MT.
We use the NLLB-200 model only for this direction as some common Zh characters are missing in the target vocabulary.

\subsection{Model training}
\subsubsection{\hok~HuBERT units}
To encode En target speech, we use the multilingual HuBERT model, the k-means quantizer and the unit vocoder released from~\citet{lee2021textless}.
Below we focus on how we build \hok~units and the corresponding unit-based speech normalizer and unit vocoder.

We train a \hok~HuBERT model using the combination of 10k-hr Mandarin speech from WenetSpeech~\citep{zhang2022wenetspeech} and 2k-hr \hok~speech from the combination of \hok~dramas, TAT and 600-hr of \hok~speech with various accents in addition to Taiwanese \hok, licensed from SpeechOcean\footnote{\url{https://en.speechocean.com/}}.
When modeling \hok~speech as discrete units, we empirically find that combining Mandarin with \hok~speech during HuBERT training allows the units to better capture the tones and produce higher-quality speech output in the unit-to-waveform conversion stage.

The HuBERT model is of the \texttt{BASE} architecture and pre-trained for three iterations following~\citet{hsu2021hubert,lakhotia2021generative}.
In the beginning of each iteration, we randomly sample $300$-hr Mandarin and \hok~speech, respectively, for k-means clustering, and apply temperature sampling to balance the amount of speech from the two languages during training.
We use $T=20$, and the probability of sampling from a language $l$ is $\tilde{p_l} = \frac{p_l^{\frac{1}{T}}}{\sum_{i}p_i^{\frac{1}{T}}}$, where $p_i = \frac{n_i}{\sum_{j}n_j}$, and $n_i$ is the number of samples from a language.
No extra language information is required during pre-training. 
In each iteration, model weights are randomly initialized and optimized for 400k steps.
We use $K=2500$ with features from the 12-th layer of the model from the third iteration for extracting \hok~units.

The \hok~speech normalizer is trained on 2-hr speech from TAT. We select speaker \textit{THF022} as the reference speaker, i.e.~the normalization target, and create speech pairs by sampling from other speakers reading the same content in TAT.
We use mask probability of 0.5, mask channel probability of 0.25 and learning rate of $3\times10^{-5}$ and train for 25k updates.
Finally, the \hok~unit-based HiFi-GAN vocoder is trained on the TTS subset of the TAT dataset, which contains a total of 36 hours of clean speech from two male and two female speakers.
We follow the training procedure in~\citet{lee2022direct} and train for 500k updates with the weight on the MSE loss for unit duration prediction set to 1.0.

\subsubsection{Wav2vec 2.0 encoder}
\label{sec:speech_encoder_training}
We pre-train the English wav2vec~2.0 encoder~\citep{baevski2020wav2vec} with the Libri-light corpus~\citep{kahn2020libri}, which contains around~54k hours of read speech audio.
The Conformer-based encoder follows the \texttt{LARGE} configuration used in~\citet{popuri2022enhanced}.
We train the encoder with a batch size of 2.1-hr for 1M updates, with~32k warmup steps and a peak learning rate of $5\times10^{-4}$.
For masking, we sample a probability of 0.065 of all time-steps to be starting indices and mask the subsequent~10 time steps.
For the \hok~wav2vec~2.0 encoder, we pre-train it with 30k-hr \hok~drama data using the same hyper-parameters as the English wav2vec~2.0 encoder.

\subsubsection{Single-pass decoding S2UT}
The \hok~unit mBART is trained with 30k-hr \hok~dramas and 10k-hr Mandarin data from WenetSpeech. The model is trained on 64 GPUs with a batch size of 3072 units.
We use Adam with a learning rate of $3\times10^{-4}$ and 10k warmup steps. The model is trained with 500K updates with dropout 0.1.
We use the En unit mBART released by~\citet{popuri2022enhanced} for training \hoken~models.

With the pre-trained wav2vec 2.0 encoder and the unit mBART decoder, we follow the best finetuning strategy in~\citet{popuri2022enhanced}, where the whole encoder and the LayerNorm and both
encoder and self attention in the decoder are finetuned with the parallel S2ST data.
The models are trained on 32 GPUs with a batch size of 160k tokens. 
We used 0.1 dropout for all models and 0.2 LayerDrop~\citep{fan2019reducing}.
The models are trained using Adam optimizer with $3\times10^{-4}$ learning rate, 10k warmup steps an 50k maximum updates.

\subsubsection{Two-pass decoding S2UT: UnitY}
\label{sec:two_pass_decoding_training}
The text mBART model is pre-trained on the combination of Mandarin and English text data from CC-100~\citep{conneau2019unsupervised}, Newscrawl~\citep{akhbardeh-etal-2021-findings}, Leipzig Corpora~\citep{goldhahn2012building}, NewsCommentary~\citep{TIEDEMANN12.463}.
There are 2B English sentences and 230M Mandarin sentences.
We learn BPE of size 65536 jointly on both languages and apply temperature sampling with $\frac{1}{T}=0.7$ during training.

We combine the pre-trained wav2vec 2.0 encoder, the text mBART decoder, and two randomly initialized Transformer layers for the text encoder and the unit decoder, respectively, to build the UnitY model.
We train our two-pass models on 16 GPUs with a batch size of 120k tokens, dropout 0.1 for all models except for the human annotated data only setup where we use dropout 0.3.
We use LayerDrop~\cite{fan2019reducing} 0.1 and label smoothing 0.1, and train the model with a learning rate of $5\times10^{-4}$, 2k warmup steps, and a maximum update of 50k steps.
The weight on the auxiliary loss from the text decoder is set to 8.0.



\subsection{Baselines}
We build two-stage and three-stage cascaded baseline systems for both \enhokbi~directions.
The two-stage cascaded system consists of a source speech (En or \hok) to target text (Mandarin or En) end-to-end S2T model and a target text to target speech unit T2U model (T2UT in the case of Zh$\rightarrow$\hok).
The three-stage cascaded system further breaks down the En$\rightarrow$Zh S2T model into En ASR followed by En$\rightarrow$Zh MT, and the \hok$\rightarrow$En S2T model is split into a \hok$\rightarrow$Zh S2T step and a Zh$\rightarrow$En MT step.

All the speech encoders for the S2T models are initialized with Conformer-based wav2vec 2.0 (Sec.~\ref{sec:speech_encoder_training}),
and the text decoders are initialized with the text mBART models (Sec.~\ref{sec:two_pass_decoding_training}).
We use the open-sourced En ASR model\footnote{\label{fn:en_asr} \url{https://huggingface.co/facebook/wav2vec2-large-960h-lv60-self}} built with the combination of wav2vec 2.0 pre-training and self-training using LibriSpeech and Libri-Light data.
We use the \enzhbi~MT models, the En T2U model and the Zh$\rightarrow$\hok~T2UT model described in Sec.~\ref{sec:weakly_supervised_data} for building the cascaded systems.



\subsection{Evaluation}
\label{sec:evaluation}
To evaluate the translation quality, we compute ASR-BLEU on the TAT-S2ST evaluation set (Sec.~\ref{sec:tat_s2st_eval_set}) by applying ASR on the generated speech and computing 4-gram BLEU against the reference text using \textsc{SacreBLEU}~\citep{post2018call}.
We use an open-sourced En ASR model\cref{fn:en_asr} when evaluating \hoken~systems.
For \enhok~systems, we build an ASR model for transcribing \hok~speech into \tailo.
The \hok~ASR is pre-trained on 10k-hr Mandarin speech from WenetSpeech and 2k-hr \hok~speech, which is a combination of TAT (480hr), \hok~dramas (1k-hr) and SpeechOcean (597-hr), with Conformer wave2vec 2.0 \texttt{LARGE} model.
We then finetuned the model with CTC loss on 480-hr \hok~speech and \tailo~scripts from TAT~\citep{liao2020formosa}, with each \tailo~syllable split into initial and final with tone as the finetuning target.
To further improve the ASR accuracy, we apply another round of self-training by generating pseudo labels on the same set of \hok~speech used in speech encoder pre-training.
The resulting \hok~ASR model achieves 9.1\% syllable error rate (SER) on the TAT-Vol1-test-lavalier set. To evaluate \enhok~translation quality, we compute syllable-level ASR-BLEU.

\section{Results}
\begin{table*}[ht!]
  \caption{Dev / test ASR-BLEU on TAT-S2ST dataset. ($^\ast$: synthetic \hok~speech is generated by applying unit vocoder on the normalized units extracted from the ground truth \hok~speech in TAT-S2ST, while synthetic En speech is generated by applying En T2U followed by the unit vocoder on the ground truth En text.)}
  \label{tab:one_vs_two}
  \centering
{\resizebox{1.0\textwidth}{!}{
  \begin{tabular}{c|l|cc|cc|cc|cc}
    \hline
\multicolumn{2}{c|}{} & \multicolumn{4}{c|}{\enhok} & \multicolumn{4}{c}{\hoken} \\
\multicolumn{2}{c|}{} & \multicolumn{2}{c|}{Training data} & \multicolumn{2}{c|}{ASR-BLEU} & \multicolumn{2}{c|}{Training data} & \multicolumn{2}{c}{ASR-BLEU} \\
\multirow{2}{*}{ID} & \multirow{2}{*}{Model} & Human & Weakly & \multirow{2}{*}{Dev} & \multirow{2}{*}{Test} & Human & Weakly & \multirow{2}{*}{Dev} & \multirow{2}{*}{Test} \\ 
& & (35-hr) & (1.5k-hr) & & & (61.4-hr) & (8k-hr) & & \\
\hline
\multicolumn{10}{l}{\textbf {Cascaded systems:}} \\
1 & Three-stage & \cmark & \cmark & 7.5 & 6.8 & \cmark & \cmark & 9.9 & 8.8  \\
2 & Two-stage & \cmark & \cmark & 7.1 & 6.6 & \cmark & \cmark & 12.5 & 10.5 \\
\hline
\multicolumn{10}{l}{\textbf {Single-stage S2UT systems:}} \\
3 & Single-pass decoding & \cmark & \xmark & 0.1 & 0.1 & \cmark & \xmark & 0.1 & 0.1 \\
4 & Single-pass decoding & \cmark & \cmark & 6.6 & 6.0 & \cmark & \cmark & 8.8 & 8.1 \\
5 & Two-pass decoding (UnitY) & \cmark & \xmark & 0.9 & 0.4 & \cmark & \xmark & 4.2 & 3.8 \\
6 & Two-pass decoding (UnitY) & \cmark & \cmark & \textbf{7.8} & \textbf{7.3} & \cmark & \cmark & \textbf{13.6} & \textbf{12.5} \\
\hline
7 & Synthetic target$^\ast$ & \xmark & \xmark & 55.5 & 53.4 & \xmark & \xmark & 76.2 & 78.5 \\
  \end{tabular}
}}
\vspace{-0.3cm}
\end{table*}

\subsection{Single-pass vs. two-pass decoding}


We first study the model architecture choice in both \enhokbi~directions. Table~\ref{tab:one_vs_two} summarizes the results.
We include ASR-BLEU from the target reference speech as a indication of the effect from the unit vocoder and the ASR errors (row 7).
We start from training on human annotated data, and it results in very low BLEU score in both directions (row 3, 5), indicating that pre-training, including wav2vec 2.0 and unit or text mBART, is not enough for building a S2ST system under low-resource for distant language pairs.
With extra supervision from text, the UnitY model works slightly better than Single-pass S2UT by 3.7 BLEU in \hoken~given 61.4-hr of S2ST data (row 3 vs.~5).

We then combine the human annotated data with weakly supervised data.
Both systems achieve significant gain (5.9-8.7 BLEU) in both directions, indicating the effectiveness of combining self-supervised pre-training and data augmentation with weakly supervised data in low-resource S2ST for a distant language pair. 
In addition, we find that UnitY outperforms single-pass S2UT in both directions (row 4 vs.~6).
In \enhok, the former achieves 1.3 BLEU higher than the latter, showing that having additional auxiliary loss based on text from a similar language can be effective in improving S2ST performance for translation into an unwritten language.
In \hoken, we obtain 4.4 BLEU gain from UnitY.
The larger impact from the additional text supervision may be due to the fact that the target text and speech are of the same language, or the larger amount of training data available for \hoken.
As the focus of this work is to present a data creation and model training strategy, we leave the investigation to future work.

For the cascaded baselines, the two-stage system works better than or close to three-stage system in both \enhokbi~directions (row 1 vs.~2). Our best performing one-stage system further outperforms the two-stage cascaded system (0.7 BLEU for \enhok~and 4.4 BLEU for \hoken, row 2 vs.~6), showing the advantage of one-stage joint training of the text and speech generation steps.

\subsection{Mined data}
In this section, we study how to leverage mined \hoken~S2T data and \enhokbi~S2ST data.

\subsubsection{Leveraging mined \enhokbi~S2ST in \enhok}


\begin{table}[t!]
  \caption{Results of \enhok~models trained with mined \enhokbi~S2ST data. We report dev / test ASR-BLEU on TAT-S2ST dataset.}
  \label{tab:mined_hoken_s2st}
  \centering
{\resizebox{0.5\textwidth}{!}{
  \begin{tabular}{c|l|ccc|cc}
    \hline
 &  & \multicolumn{3}{c|}{Training data} & \multicolumn{2}{c}{ASR-BLEU} \\
\multirow{2}{*}{ID} & \multirow{2}{*}{Model} & Human & Weakly & Mined & \multirow{2}{*}{Dev} & \multirow{2}{*}{Test} \\
 & & (35-hr) & (1.5k-hr) & (197-hr) & \\
\hline
 3 &   & \cmark & \xmark & \xmark & 0.1 & 0.1 \\
 8 & Single-pass & \cmark & \xmark & \cmark & 0.1 & 0.1 \\
 4 & decoding & \cmark & \cmark & \xmark & 6.6 & 6.0 \\
 9 &  & \cmark & \cmark & \cmark & 6.7 & 6.0 \\
\hline 
 5 &  & \cmark & \xmark & \xmark & 0.9 & 0.4 \\
 10 & Two-pass & \cmark & \xmark & \cmark & 5.7 & 4.9 \\
 6 & (UnitY) & \cmark & \cmark & \xmark & 7.8 & 7.3 \\
 11 & & \cmark & \cmark & \cmark & 8.0 & 7.5 \\
\hline 
  \end{tabular}
}}
\vspace{-0.3cm}
\end{table}

In Table~\ref{tab:mined_hoken_s2st}, we show the results of leveraging the mined \enhokbi~S2ST data in \enhok~direction.
In order to train the UnitY model, we apply \hok$\rightarrow$Zh S2T to generate pseudo-labeled Mandarin text for the mined \hok~speech as the auxiliary task target.

We first train both one-stage models with mined data and the human annotated data.
While the single-pass decoding S2UT model still yields very low BLEU score (row 8), the UnitY model achieves 4.5 BLEU improvement with the extra 197-hr of mined S2ST data (row 5 vs.~10), showing that noisy Mandarin text generated from pseudo-labeling still provides useful signals in model training.
We then further combine with weakly supervised data but do not see significant gain with the additional mined data (row 4 vs.~9, 6 vs.~11).
Note that the size of mined data is only 13\% of the total amount of weakly supervised data we have.
As discussed in Sec.~\ref{sec:exp_setup_mined_data}, the limited amount of mined data available is mainly due to the domain mismatch issue.
In the future, we plan to explore mined data from more similar domains and aim to increase the amount of data for better S2ST performance.

\subsubsection{Leveraging mined \hoken~ST in \hoken~direction}
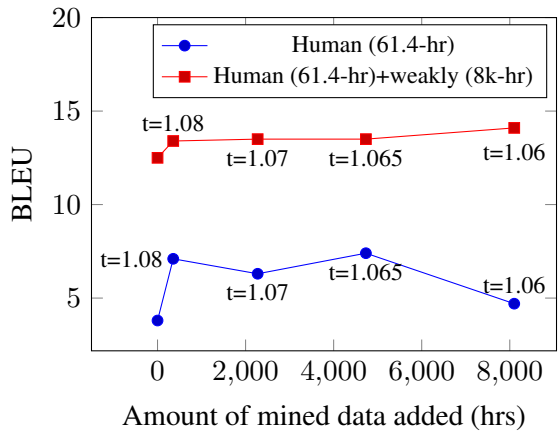
\begin{figure}
    \centering
    \begin{tikzpicture}
\begin{axis}[
  xlabel=Amount of mined data added (hrs),
  ylabel=BLEU,
  ymax=20,
  xmin=-1500,
  width=1.0\columnwidth,
  height=6cm
]
\addplot table [y=testbleu, x=hours]{mined_threshold_human.dat};
\node [above, font=\small] at (axis cs:  8101, 4.7 ) {t=1.06};
\node [below, font=\small] at (axis cs:  4732, 7.4 ) {t=1.065};
\node [below, font=\small] at (axis cs:  2274, 6.3 ) {t=1.07};
\node [left, font=\small] at (axis cs:  356, 7.1 ) {t=1.08};
\addplot table [y=testbleu, x=hours]{mined_threshold_human_weakly.dat};
\node [below, font=\small] at (axis cs:  8101, 13.8 ) {t=1.06};
\node [below, font=\small] at (axis cs:  4732, 13.5 ) {t=1.065};
\node [below, font=\small] at (axis cs:  2274, 13.5 ) {t=1.07};
\node [above, font=\small] at (axis cs:  356, 13.4 ) {t=1.08};
\addlegendentry[font=\small]{Human (61.4-hr)}
\addlegendentry[font=\small]{Human (61.4-hr)+weakly (8k-hr)}
\end{axis}
\end{tikzpicture}
    \caption{BLEU scores on TAT-S2ST \hoken~test set from UnitY models trained with mined data filtered at different thresholds (t) for the similarity score.}
    \label{fig:hoken_mined_threshold}
\end{figure}

We convert the mined \hoken~S2T data to S2ST data with the En T2U model and train UnitY models with the combination of human annotated data and optionally the 8k-hr weakly supervised data to examine the effect of mined data on model performance.
Fig.~\ref{fig:hoken_mined_threshold} shows the ASR-BLEU scores on the TAT-S2ST test set with respect to different thresholds on the similarity scores of the mined pairs.

We see that adding 4.7k-hr mined \hoken~S2T data ($t=1.065$) is helpful to improve the model quality by 3.6 BLEU when only human annotated data is available.
With 8.1k-hr mined data ($t=1.06$), the BLEU gain drops to 0.9 BLEU.
In addition, there is a 7.8 BLEU gap lower towards the UnitY model trained with human annotated data and 8k-hr of weakly supervised data (Table~\ref{tab:one_vs_two} row 6).
As the \hok~speech for both weakly supervised data and mined data come from the same \hok~dramas dataset, the gap implies that pseudo-labeling is a generally effective data augmentation technique for low-resource scenarios, while the quality of the mined data is constrained by the content of the data available for mining. 
However, combining all three types of data together is still beneficial.
We obtain 1.6 BLEU gain by adding 8.1k-hr mined data to the combination of human annotated and weakly supervised data.

\section{Conclusions}
We present the first \enhokbi~S2ST systems, where \hok~is an oral language that does not have standard and widely adopted text writing systems, i.e.~an unwritten language.
To tackle the challenges of speech translation for unwritten languages and the lack of parallel training data, we present an end-to-end study.
First, we explore three options of training data creation including human annotation, weakly supervised data from pseudo-labeling and data mining.
Second, we investigate two modeling choices including direct speech-to-unit translation with a single speech unit decoder and two-pass decoding that leverages extra supervision from target text.
Experimental results show that leveraging a similar high-resource written language (Mandarin in the case of \hok) is effective in both the data creation process as well as model training.
Finally, we release the benchmark dataset and ASR evaluation model to facilitate research in this field.
In the future, we aim to expand study and establish an S2ST model building strategy that works for a diverse set of unwritten languages.

\section{Acknowledgements}
We would like to thank Koklioong Loa for consulting on Taiwanese \hok; Eric Lam, Kai-Wei Chang and Hung-yi Lee from National Taiwan University for providing \hok~drama data; Janice Lam for writing and fact-checking the information about \hok~language; Brian Bui and Carleigh Wood for coordinating the data annotation effort; Ilia Kulikov for setting up the evaluation script.

\bibliography{refs}
\bibliographystyle{acl_natbib}


\end{document}